%% file: main.tex
\documentclass[letterpaper]{article}
\usepackage{proceed}
\usepackage[margin=1in]{geometry}

\usepackage{graphics} 
\usepackage{mathptmx} 
\usepackage{times} 
\usepackage{amsmath} 
\usepackage{amssymb}  
\usepackage[separate-uncertainty=true, per-mode=symbol]{siunitx} 
\usepackage[capitalize]{cleveref}
\usepackage{booktabs}
\usepackage{url} 
\usepackage{standalone} 
\usepackage{tikz}  
\usepackage{pgfplots} 
\usepackage[noend]{algcompatible}
\usepackage{algorithm}
\usepackage{xcolor}

\newcommand{\RETURN}{\textbf{return }}
\usepackage[normalem]{ulem} 

\usepgfplotslibrary{groupplots}
\usetikzlibrary{arrows.meta}
\usetikzlibrary{arrows}
\usetikzlibrary{positioning}
\usetikzlibrary{shapes}
\usetikzlibrary{backgrounds}
\tikzset{>=stealth'}
\pgfplotsset{compat=newest}
\pgfplotsset{every axis legend/.append style={%
		cells={anchor=west},
		label style={font=\Huge}}
}
\usepgfplotslibrary{polar}

\usepackage[backend=bibtex,natbib,style=authoryear,uniquename=init,giveninits=true, maxcitenames=1, maxbibnames=100]{biblatex}
\AtEveryBibitem{
	\ifentrytype{inproceedings}{
		\clearlist{address}
		\clearlist{publisher}
		\clearname{editor}
		\clearlist{organization}
		\clearfield{url}  
		\clearfield{doi}  
		\clearfield{pages}  
		\clearlist{location}
	}{}
}

\AtBeginBibliography{\small}

\title{Reinforcement Learning with Probabilistic Guarantees \\ for Autonomous Driving }

\author{
	{\bf Maxime Bouton$^1$, Jesper Karlsson$^2$, Alireza Nakhaei$^3$, Kikuo Fujimura$^3$, Mykel J. Kochenderfer$^1$, Jana Tumova$^2$} \\
    $^1$Stanford University, Stanford, CA, \{boutonm, mykel\}@stanford.edu \\
    $^2${KTH} Royal Institute of Technology, Stockholm, Sweden \{jeskarl, tumova\}@kth.se \\
    $^3${H}onda {R}esearch {I}nstitute, Mountain View, CA, \{anakhaei, kfujimura\}@honda-ri.com    
}

\begin{document}

\maketitle

\begin{abstract}

Designing reliable decision strategies for autonomous urban driving is challenging. 
Reinforcement learning (RL) has been used to automatically derive suitable behavior in uncertain environments, but it does not provide any guarantee on the performance of the resulting policy. 
We propose a generic approach to enforce probabilistic guarantees on an RL agent. 
An exploration strategy is derived prior to training that constrains the agent to choose among actions that satisfy a desired probabilistic specification expressed with linear temporal logic (LTL). 
Reducing the search space to policies satisfying the LTL formula  helps training and simplifies reward design. 
This paper outlines a case study of an intersection scenario involving multiple traffic participants. The resulting policy outperforms a rule-based heuristic approach in terms of efficiency while exhibiting strong guarantees on safety.

\end{abstract}

\section{Introduction}

\begin{figure}[ht!]
	\centering
	\includegraphics[width=\columnwidth]{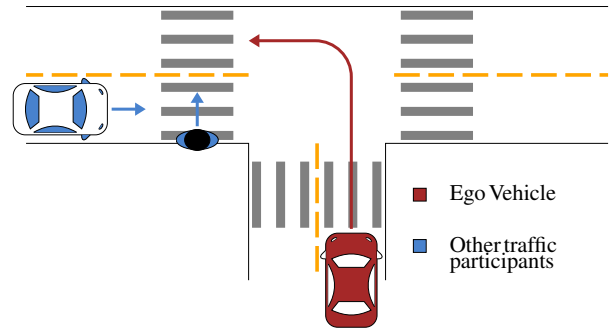}
	\caption{The ego vehicle must achieve a left turn in an unsignalized intersection involving another vehicle and a pedestrian interacting with each other.}
	\label{fig:left-turn}	
\end{figure}

It is often desirable to prove that a decision policy is safe before testing it in the real world. 
Formal methods provide tools to verify a given policy with high confidence but often rely on very strong modeling assumptions and are not suitable to derive optimal strategies. 
In contrast, reinforcement learning algorithms address this issue by optimizing decision policies without an explicit model of the environment. 
An agent interacts with the environment (through simulation, collected data or directly with the real world) and optimizes an expected long term return modeled by a reward function. 
This function reflects a high level objective by attributing a positive reward or a penalty on the immediate outcome of an action. 
Although the resulting strategy might achieve a high expected return, no guarantees can be made about its performance. 
Moreover, in many applications, the reward function might be difficult to design, causing the resulting behavior to be misaligned with the desired objective. 

Previous work has investigated combining formal methods and reinforcement learning using a monitoring system enforcing the learning agent to satisfy a safety property~\citep{fulton2018, alshiekh2018}. 
However, designing such a monitoring system can be challenging. It may involve conservative assumptions about the environment and rely on deterministic models. 
When uncertainty is very high, imposing hard safety constraints on the learning agent can result in every action resulting in a violation of the constraints. 
Often in applications such as autonomous driving, it is necessary to accept some level of risk in order to achieve the objectives of the system.

We propose a generic approach to train an agent in an uncertain environment while enforcing a minimum probability of satisfying a specification expressed using Linear Temporal Logic (LTL) \citep{baier2008}. 
Natural language can be converted to LTL formulas \citep{baier2008} which makes it an interesting framework to mathematically express specifications derived by regulators, policy makers or governmental agencies. 
By modeling the environment as a Markov decision process (MDP), we identify at each state, the set of actions satisfying the specification through model checking techniques. 
We then show how to constrain a reinforcement learning agent to only choose among the acceptable actions. In contrast with previous work \citep{fulton2018,alshiekh2018,mukadam2017}, our framework  handles probabilistic specifications that accommodate less conservative policies. At the end of the training, the policy is guaranteed to satisfy the LTL formula with a user-defined confidence level. Finally, we discuss applications to unsignalized intersection scenarios involving a car and a pedestrian as illustrated in~\cref{fig:left-turn}.

\section{Related Work}

Previous approaches attempted to generate safe policies for urban driving by tuning a penalty for collision in the reward function~\citep{bouton2017}. The reward function can be used to balance conflicting objectives such as safety and efficiency by varying the cost associated with collisions. Metrics, such as time to reach the goal and collision rate, are then used to choose a suitable operating point. Given a reward function, the policy is empirically evaluated through simulations. Empirical validation is hard because it requires an enormous number of simulations to cover every possible situation that might cause unsafe behavior.

Tools from formal methods can be used to verify properties of a system with high confidence. A possible approach involves verifying the policy after the planning or learning phase~\citep{katz2017} but requires restarting the planning if an issue is found. Another method is to derive a  policy from a boolean requirement rather than a reward function~\citep{wang2018}. However, due to tractability issues, they are limited to small discretized problems making them unsuitable for practical use in complex environments.

Many approaches to derive safe policies in a reinforcement learning context consist of constraining the policy space to known safe actions~\citep{fulton2018, garcia2015, jansen2018}. A reinforcement learning algorithm to derive provably safe policies has been demonstrated for hybrid systems in~\citep{fulton2018}. The authors demonstrate the use of a model monitor that allows for taking exploratory actions when the execution environment is different from the training or planning environment. This approach was successfully applied to a simple autonomous driving scenario, but it is unclear how it would scale to more complex scenarios and it does not handle LTL semantics. Existing approaches involving LTL formulas and monitoring RL agent rely on a deterministic abstraction of the environment~\citep{alshiekh2018}. . Others have been relying on ad-hoc knowledge to design a rule-based system that would prevent an RL agent to take unsafe actions~\citep{mukadam2017}. Another approach consists of learning an MDP model from observation and solve a model checking problem over the learnt model~\cite{jansen2018}. Similarly as in our approach, the result from model checking is then used to monitor a reinforcement learning agent. \citeauthor{jansen2018} uses a finite horizon model checking algorithm to constrain the RL agent. In this work we use an infinite horizon method. Our approach can handle LTL formulas and probabilistic specifications and is not tailored for a specific scenario.


\section{Background}

This section introduces the common mathematical framework for formulating and solving decision problems in stochastic environments.

\subsection{Markov Decision Processes}

A Markov decision process (MDP) is defined by the tuple $(\mathcal{S}, \mathcal{A}, T, R)$, where $\mathcal{S}$ is a state space (possibly continuous), $\mathcal{A}$ is an action space, $T$ is a state transition function, and $R$ is a reward function \citep{Bellman1957,dmu}. At time $t$, an agent chooses  action $a_t\in\mathcal{A}$ based on observing state $s_t\in\mathcal{S}$ and receives reward
 $r_t=R(s_t, a_t)$. 
The environment transitions from $s_t$ to $s_{t+1}$ with  probability $\Pr(s_{t+1} \mid s_t, a_t) = T(s_{t+1}, s_t, a_t)$. The action selection strategy is called a policy and is defined as a mapping from state to action, $\pi: \mathcal{S} \mapsto \mathcal{A}$. Throughout the paper, we focus on problems with a finite number of actions.

When performing model checking, a convenient approach is to label the states of the MDP and express the property we wish to verify in terms of these labels. 
The labels are atomic proposition that evaluates to true or false at a given state. We define a \textit{labelled Markov decision process} by the tuple $(\mathcal{S}, \mathcal{A}, T, R, \Pi, L)$ where $\Pi$ is a finite set of atomic propositions, and $L$ a mapping, $L:\mathcal{S} \mapsto 2^{\Pi}$, giving the set of atomic propositions satisfied at a given state. The other elements of the tuple are the same as in the previous definition.


\subsection{Reinforcement Learning}

In reinforcement learning (RL), the environment is modeled as an MDP with unknown transition and reward models \citep{KLMSurvey}. The agent observes the state of the environment and receives a reward at every time step. While gathering experience, the agent updates its policy to maximize its accumulated reward. This paradigm does not require any explicit model of the environment and can optimize decision policies in uncertain environments. However, RL approaches do not provide strong guarantees on the performance of the resulting policy. 

This work focuses on model-free reinforcement learning algorithms such as Q-learning \citep{watkins1989}, where the policy is represented by a state action value function (Q function). The optimal Q function, satisfies the Bellman equation:
\begin{equation}
	Q^*(s, a) = \mathbb{E}_{s'}[R(s, a) + \gamma \max_{a'}Q^*(s', a')]
	\label{eq:bellman}
\end{equation}

When the state space is large, we rely on parametric representation of the Q function such as deep neural networks. A loss function is formulated from \cref{eq:bellman} and the parameters are updated using gradient descent \citep{dmu}. 

Although Q-learning is a model-free algorithm, the policy is often trained in a simulation environment that relies on models. The model-free assumption is still useful because it does not require any specific form for the model apart from the ability to generate the next state of the environment given the previous state and an action. When a model is not available, sampled data can be used. 

\subsection{Linear Temporal Logic}

To express the probabilistic specification, we use Linear Temporal Logic (LTL). LTL is an extension to propositional logic taking into account temporal modalities. An LTL formula is built of atomic propositions defined over the states of the MDP (using the labels) according to the following grammar:
\begin{equation}
  \phi ::= a \mid \phi_1 \land \phi_2 \mid \phi_1 \lor \phi_2 \mid \lnot\phi \mid \mathsf{G} \phi \mid \mathsf{F} \phi \mid \phi_1 \mathsf{U} \phi_2 \mid \mathsf{X} \phi
  \label{eq:ltl}
\end{equation}
where $a$ is an atomic proposition, $\phi$, $\phi_1$, and $\phi_2$ are LTL formulas, $\lnot$ (negation), $\land$ (conjunction), and $\lor$ (disjunction) are logical operators, and $\mathsf{G}$ (globally), $\mathsf{F}$ (eventually), $\mathsf{U}$ (until), and $\mathsf{X}$ (next) are temporal operators~\citep{baier2008}. 
The satisfaction of an LTL formula is evaluated on an infinite sequence of states in the MDP. Given an MDP and a policy $\pi$, $\Pr^\pi(s \models \phi)$ denotes the probability that the LTL formula $\phi$ will be satisfied by the sequence of states obtained when executing $\pi$ starting from state $s$.  

\section{Proposed Approach}

This section introduces our two-step approach to derive a decision making policy with probabilistic guarantees. Actions or policies that satisfy the desired specification are referred to as acceptable actions or policies. We first describe how to identify the set of acceptable actions. We then demonstrate how to derive acceptable policies in a reinforcement learning framework by constraining the exploration strategy of the agent. 

\subsection{Identifying Acceptable Actions}\label{sec:model_checking}

A model checking approach is used to identify at each state, the set of actions that satisfy a desired specification. The specification is expressed using an LTL formula and a threshold on the probability of satisfaction.  For example one could specify the formula $\mathsf{G} \lnot \verb|collision|$ with a threshold of \num{0.99}, which states that the probability that a collision never occurs should be greater than \num{0.99}. The threshold can be interpreted as an acceptable level of risk.

We address the general problem of identifying which actions can be taken to ensure that the probability of satisfaction of the LTL formula is above a threshold. We compute the probability of satisfying a given LTL property $\phi$ at a given state action pair. The satisfaction of $\phi$ depends on the policy followed by the agent. As a consequence, we compute the maximum probability of satisfying $\phi$ when starting in a state $s$ and taking an action $a$: $\Pr^{\max}(s, a \models \phi)$ where the maximum is over the possible policies. Existing model checking algorithms address the problem of computing $\Pr^{\max}(s \models \phi)$ \citep{lahijanian2011,lahijanian2010motion, baier2008}. Most of these techniques can be adapted to derive the desired state action quantity using the following relation between the two quantities: $\Pr^{\max}(s \models \phi)= \max_a \Pr^{\max}(s, a \models \phi)$. Our work relies on a value iteration algorithm. 

For any LTL formula, computing the maximum probability of satisfaction for every state of an MDP can be reduced to a reachability problem, \textit{i.e.} computing the maximum probability of reaching a set of states $\mathcal{B}$~\citep{baier2008}. The set $\mathcal{B}$ is often expressed in terms of the states labels. This problem is then solved using value iteration. \Cref{alg:value_iteration} presents an adapted version of the value iteration algorithm to solve reachability problems where line \num{6} can be interpreted as a Bellman update.

\begin{algorithm}
	\caption{Value Iteration algorithm to compute the maximum probability of reaching a set of state}
	\begin{algorithmic}[1]
		\STATE \textbf{input: } a labelled MDP, a set of states $\mathcal{B}$
		\STATE Initialize $P^{(0)}(s, a)$ to 1 if $s \in \mathcal{B}$, 0 otherwise
		\STATE $k \leftarrow 0$ 
		\REPEAT
        \STATE For all state action pair $s,\, a$:
		\STATE $P^{(k+1)}(s, a) \leftarrow \sum_{s'\in \mathcal{S}} \Pr(s'\mid s, a)\max_a P^{(k)}(s', a)$
		\STATE $k \leftarrow k+1$
		\UNTIL{convergence}
	\end{algorithmic}
    \label{alg:value_iteration}
\end{algorithm}

Once the quantity $\Pr^{\max}(s, a \models \phi)$ is computed for each state action pair, we can identify acceptable actions that the agent can take without violating the specification. Given the desired threshold $\lambda$, the set of acceptable actions is given by: 
\begin{equation}
Act(s) = \{a \mid \Pr\,^{\max}(s, a \models \phi) > \lambda \} \text{ .}
\label{eq:act}
\end{equation} 

We can extract the policy that ensures the maximum level of satisfaction: $\pi^*(s) = \arg\max_a \Pr^{\max}(s, a \models \phi)$. The quantity $\Pr^{\max}(s, a \models \phi)$ can be interpreted the following way: the maximum probability of satisfying $\phi$ when starting in state $s$ and taking action $a$ and then following $\pi^*$ for the next time steps. It is analogous to the state action Q function in reinforcement learning.

The complexity of value iteration is polynomial in the number of states and actions, and it can be parallelized. For specific subclasses of LTL formulas, the mapping to a reachability problem is not necessary, and it can be directly computed using variants of \cref{alg:value_iteration} that have the same convergence property and the same complexity~\citep{lahijanian2011}. \Cref{alg:value_iteration} requires a model of the transition distribution of the MDP. For the result of the model checking approach to hold for real world problem, it is important that the model is conservative.

In many problems, the agent interacts with a naturally continuous environment. To address this issue, the state space can be discretized in a grid. Using \cref{alg:value_iteration}, the probability of satisfying the desired property is computed for each state in the grid. Generalizing to other states can be done via multi-linear interpolation, nearest neighbors, or other forms of approximation \citep{dmu}. We can still draw strong probability guarantees in the continuous domain by including the error introduced by discretization in the threshold. Model-free problems can also be addressed by the approximation methods via sampling. Scaling the current algorithm to large state spaces and model free setting is left as future work.

\subsection{Reinforcement Learning with Probabilistic Guarantees}

The result of the model checking approach enables distinguishing between acceptable and non acceptable actions with high confidence (governed by the specified threshold). This ability can be used to monitor a reinforcement learning agent such that the probabilistic specification holds during training and during the execution of the resulting policy. 


In order to maintain the desired property during training, the exploration process of the reinforcement learning algorithm is modified. The learning agent is constrained to select an action from the set given by \cref{eq:act}. When the set is not empty, it can use any randomization scheme like $\epsilon$-greedy, the agent selects a random action in the set with a probability $\epsilon$ and selects the action that maximizes the state action Q function otherwise (still among the set). This exploration strategy does not prevent the agent visiting states where no actions are available. In this case, the exploration policy default to $\pi^*$ which is the best possible policy to satisfy the LTL specification. A pseudo-code is provided in \cref{alg:exploration}.

\begin{algorithm}
	\caption{Modified $\epsilon$-greedy exploration policy }
	\begin{algorithmic}[1]
		\STATE \textbf{input: } a state $s$; the result from model checking: $\pi^*$, $Act$; the current $Q$ function 
        \IF{$Act(s) = \varnothing$}
        	\STATE \RETURN $\pi^*(s)$
        \ELSE 
          \STATE sample $r\sim\text{Uniform}(0, 1)$
          \IF{$ r < \epsilon $}
            \STATE \RETURN random action from $Act(s)$
          \ELSE
            \STATE \RETURN $\arg\max_{a\in Act(s)} Q(s, a)$ 
           \ENDIF
         \ENDIF
	\end{algorithmic}
    \label{alg:exploration}
\end{algorithm}

The trained policy is also constrained. The trained agent takes the acceptable actions that maximizes the Q function and following $\pi^*$ if there are no acceptable actions at the current state. 

The modified exploration strategy does not affect the convergence of the value function in the states that have a high probability of satisfying the specification. A whole part of the state space is avoided during training and the value function for these states will not be accurate. However, the policy is still reliable since the agent relies on $\pi*$ when reaching these non acceptable states. This exploration strategy can be applied to any generic reinforcement learning algorithm. \Cref{sec:application}  demonstrates an example using Deep Q learning \citep{mnih2015}. 


Safe reinforcement learning methods can be categorized into two approaches: modification of the optimality criterion or modification of the exploration process~\citep{garcia2015,fulton2018}. The proposed technique focuses on modifying the exploration strategy. Another approach consists of modifying the reward function by adding a penalty when the agent violates the LTL specification. However designing a reward signal matching the goal expressed by the LTL formula might be very challenging. Moreover, a more complex reward signal is likely to make the training less efficient. 

Our approach allows us to decouple some of the objectives of the problem, which can help simplify the reward signal. The reward function used in reinforcement learning might express a different objective than the one specified through the LTL formula. Our strategy can be interpreted as a constrained optimization problem. We first identify the space of policies where a desired LTL specification is satisfied and then find the policy in this space that maximizes the accumulated reward. The constraints (\textit{e.g.} safety) are encoded in the LTL formulas and handled using the model checking approach. Decoupling the objectives of the problem can help the training significantly as demonstrated in \cref{sec:application}. For example, if a safety constraint is handled by the model checking approach, then the RL agent will only explore safe states, and will not have to learn how to avoid unsafe regions. 

\section{Case Study}\label{sec:application}

This section considers a case study using the model checking approach and the modified reinforcement learning procedure. It provides an empirical comparison of our framework against different baselines methods.  

\subsection{Autonomous Driving at Unsignalized Intersections}

To illustrate our approach, we chose to focus on an autonomous driving scenario at an unsignalized intersection involving another car and a pedestrian as illustrated in \cref{fig:left-turn}. The agent (ego vehicle) must achieve a left-turn in the intersection safely and efficiently. The other traffic participants have uncertain behaviors and interact with each other and with the ego vehicle. The ego vehicle must anticipate their behaviors and maintain safety while trying to reach a goal position as fast as possible. The pedestrian can cross at any of the three crosswalks. The other car comes from the horizontal street, from the left or from the right, and can either turn or go straight. For this scenario, our desired probabilistic specification will be expressing safety. A maximum acceptable level of risk is imposed on the ego vehicle.

This problem can be modeled as a Markov decision process. Previous works propose modeling decision making problems in urban navigation scenario as Markov decision processes or partially observable Markov decision processes \citep{sunberg2017, bouton2017, mukadam2017, bandyopadhyay2012}. An intersection navigation problem is addressed using a tree search method, considering only cars in \citet{bouton2017}. A lane-changing scenario is addressed using RL in \citet{mukadam2017} but they use a handcrafted approach to ensure safety. To our knowledge, none of the previous approaches provide probabilistic guarantees on the resulting strategy, the performance is evaluated in simulation and tuned through the reward function. In order to scale these approaches to environments with multiple traffic participants, decomposition methods can be used \citep{bouton2018utility}. 

We consider three subproblems of various complexity based on the number of agents to avoid: a scenario with one single pedestrian, one with one single car and a scenario with both a car and a pedestrian. This last scenario allows to capture the interaction between the vehicle and the pedestrian. 

The state of the environment consists of the physical state of the three traffic participants; the ego vehicle, the human driver and the pedestrian. A physical state is described in lane relative coordinates. It consists of the longitudinal position, the longitudinal velocity, the lane, as well as the route. The route indicates the starting position of the traffic participant and its goal, it encodes vehicles intentions such as turning right. In this work we assume that the route is perfectly observable. Modeling intentions as partially observable state variables, as in \citet{bandyopadhyay2012}, is left as future work. The other car can take one of four possible routes according to the topology of the intersection: straight from the left, straight from the right, turn left, turn right. A physical state can also take the specific value $s_{absent}$ to model a potential incoming vehicle or pedestrian. 

The action space is a set of strategic maneuvers corresponding to hard braking, moderate braking, constant speed, and acceleration. They correspond to longitudinal accelerations along the given path: $\{\SI{-4}{\meter\per\second\squared},\SI{-2}{\meter\per\second\squared},\SI{0}{\meter\per\second\squared},\SI{2}{\meter\per\second\squared}\}$.

The ego vehicle moves according to a discrete time point mass model. The pedestrian follows a constant speed of \SI{1}{\meter\per\second} with a random variation uniformly drawn from $\{\SI{-1}{\meter\per\second},\SI{0}{\meter\per\second},\SI{1}{\meter\per\second}\}$. 
The other car moves according to a simple rule based policy:
\begin{itemize}
    \itemsep 0.2em
	\item always yield to the pedestrian if it is crossing
    \item give priority to other vehicles when turning left and rely on the time to collision to decide when to cross  
    \item follow the Intelligent Driver Model otherwise \citep{treiber2000}
\end{itemize}
This simple strategy outputs an acceleration, at which is added a random amount drawn uniformly from the set
$\{\SI{-1}{\meter\per\second\squared},\SI{0}{\meter\per\second\squared},\SI{1}{\meter\per\second\squared}\}$. When the other car, or the pedestrian, is in the state $s_{absent}$, there is a probability of \num{0.7} that it appears at the beginning of a  lane, or at a crosswalk, at the next time step. They appear with a random velocity, between \SI{0}{\meter\per\second} and \SI{8}{\meter\per\second}, and with a random route (turn or straight) for the car.

Formulating a joint problem with three agents allows to capture interactions. In this example, the other vehicle is interacting with the pedestrian and with the ego vehicle. It is expected that leveraging this interaction will result in a more efficient policy.

We provided the following specification: ${\phi = \lnot\verb|collision| \mathsf{U} \verb|goal|}$ with a threshold of \num{0.9999}, where $\verb|collision|$ evaluates to true when the ego vehicle collides with another traffic participant and $\verb|goal|$ is verified when the ego vehicle achieves the left turn and reaches a position \SI{20}{\meter} from the intersection. Satisfying this specification means avoiding any collisions.

\subsection{Computing a safe policy}

In order to compute a safe policy that satisfies the specified risk tolerance, we first run \cref{alg:value_iteration} to compute $\Pr^{\max}(s, a \models \phi)$. For the simple safety specification provided, the set $\mathcal{B}$ corresponds to the set of goal states. Collision states are terminal states and the probability of reaching the goal state from these states is zero. 

Before applying \cref{alg:value_iteration}, the continuous state space must be discretized. The position resolution is set to \SI{2}{\meter} and the velocity resolution is set to \SI{2}{\meter\per\second} for vehicles and \SI{1}{\meter\per\second} for pedestrians. This discretization leads to \num{204} physical states for the ego vehicle, \num{793} for the other car, \num{145} states for the pedestrian. 

When computing \cref{eq:act} in the continuous space environment, we use multi-linear interpolation to map the continuous state of the traffic participants to a state on the grid \citep{dmu}. The value at the continuous state is then approximated by a weighted sum of the value of the interpolants:
\begin{equation}
	\Pr(s, a \models \phi) \approx \sum_{s_i \in \text{Interpolants}}w_i\Pr(s_i, a \models \phi)
\end{equation}
where $w_i$ is a weight proportional to the distance between the continuous state $s$ and the interpolant $s_i$ in the grid. The multi-linear interpolation is carried independently for each traffic participant on a two dimensional grid corresponding to longitudinal position and velocity. The error introduced by this approximation can be compensated by a more conservative threshold.


\subsubsection{Train the policy with constrained exploration}

The solution from the model checking part is then used to constrain the exploration of the agent in a reinforcement learning algorithm. In this work we used the Deep Q Learning algorithm \citep{mnih2015} with prioritized experience replay \citep{schaul2016}. In Deep Q Learning the Q function is represented by a neural network enabling the algorithm to handle continuous states. To make the state variable a suitable input for the neural network we converted them to Cartesian coordinates, each vehicle is represented by four dimensions (two for the position, one for the heading, one for the velocity). Each variable is normalized. The four routes are represented by a one hot encoding. For the single pedestrian, the input dimension is eight. For the single car problem, it is a twelve dimensional vector since the possible route of the other vehicle must be considered. We used three fully connected layers of thirty-two nodes for the neural network architecture for all scenarios and the same hyperparameters as those of \citet{bouton2018utility}.




\subsection{Experiments}

The results of \num{10000} Monte Carlo simulations are reported in \cref{tab:results}. We evaluated the policies regarding two metrics: the collision rate and the average number of time steps to reach the goal position. Our approach is compared to different baselines. We adopted the following terminology:

  \textbf{Safe Random:} At each state, the agent chooses a random action among the set $Act(s)$. This policy is expected to satisfy the LTL specification while being very inefficient with respect to the reward function. We included these result as an empirical verification of the model checking algorithm.
  
  \textbf{Rule Based:} The ego vehicle follows a handcrafted rule-based policy which is similar to the one followed by other vehicles in the simulation environment. This policy does not result of any optimization procedure but provides an assessment of the difficulty of the environment.  
  
  \textbf{RL:} A policy is trained to maximize an accumulated reward function. In addition to the reward for reaching the goal, the agent is penalized for colliding with other traffic participants. This is a typical RL procedure, the policy is expected to be efficient but will not necessary satisfy the specification. It is optimizing a simple reward function attributing a reward of \num{1.0} for reaching the goal, a high penalty for colliding with another agent, and a smaller penalty for each action. The collision penalty was varied between \num{-1.0} and \num{-5.0} and the action cost between \num{-0.05} and \num{0.0}. By increasing the collision cost, the policy is more risk averse and more likely to satisfy the specification. In contrast, increasing the action cost makes it reach the goal faster, at the expense of a greater risk. 
  
  \textbf{Safe RL:} The policy resulting from training a reinforcement learning agent constrained to acceptable actions. The reward function used for training is similar as the one described above. However, since the agent is constrained to safe action only, it will never receive the penalty associated to collision. It is only optimizing for reaching the goal faster.

For the scenario involving both a car and a pedestrian we approximated the set of acceptable actions as the intersection of the sets given by solving the simpler problems of avoiding one car and one pedestrian individually. This approach is suboptimal since it does not leverage the interaction between the two traffic participants. When the intersection is empty, the ego vehicle default to a hard-braking action which is always safe in the scenario of interest. Another approach would be to identify acceptable actions in the joint problem directly. The state space of this problem is much larger and hence the computation of $\Pr^{\max}(s, a \models \phi)$ would be more expensive. 

\begin{table}[h!]
	\centering
	\caption{Performance on the different scenarios}
	\begin{tabular}{lSS}
		\toprule[1pt]
		& \text{Collision rate (\si{\percent})} & \text{Time steps} \\
        \midrule
		\multicolumn{3}{l}{\textbf{Single Pedestrian}} \\
		Safe Random   & 0.00  &  77.06\\
		Rule Based & 0.05 & 41.83 \\
		RL     & 0.63 & 22.71 \\
        Safe RL   &  0.00  & 17.24 \\
        \midrule
		\multicolumn{3}{l}{\textbf{Single Car}} \\
        Safe Random  &  0.00  &  76.28  \\
		Rule Based & 2.18 & 26.62 \\
		RL     & 0.2 & 14.8 \\
		Safe RL   &  0.00  & 22.44 \\
        \midrule
		\multicolumn{3}{l}{\textbf{Car and Pedestrian}} \\
        Safe Random &  0.00 & 87.304  \\
		Rule Based & 3.62  & 42.05 \\
		RL     & 0.96 & 22.16  \\
		Safe RL  &  0.00  & 28.47 \\
		\bottomrule[1pt]
	\end{tabular}
	\label{tab:results}
\end{table}

\begin{figure}[ht!]
	\centering
    \input{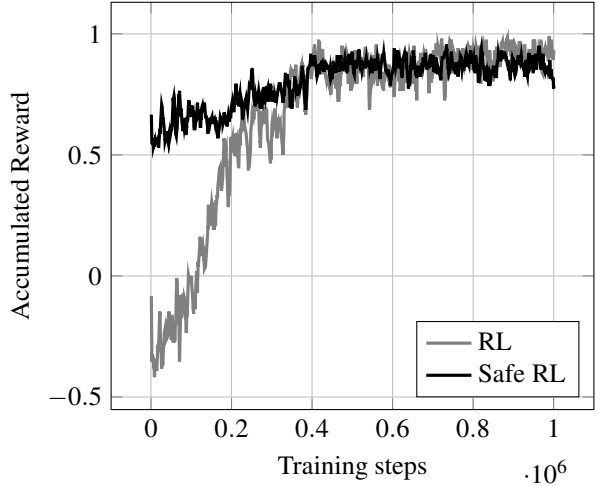}
    \caption{Evolution of the accumulated reward during training on the scenario with a car and a pedestrian. Both policies are trained with the same reward function: \num{-1.0} for collision, \num{1.0} for reaching the goal and no action~cost.}
    \label{fig:training}
\end{figure}

The training performance of the safe RL and regular RL approaches is reported in \Cref{fig:training}. The policies are trained with the same reward function. One million steps in the environment were used to ensure the convergence for both policies.

\begin{figure}[ht!]
	\centering
    \input{pareto.tex}
    \caption{Pareto frontiers of the Safe RL technique and regular RL on the scenario with one car and one pedestrian. Different behavior are generated by changing the action cost and the collision cost in the reward function.}
    \label{fig:pareto}
\end{figure}
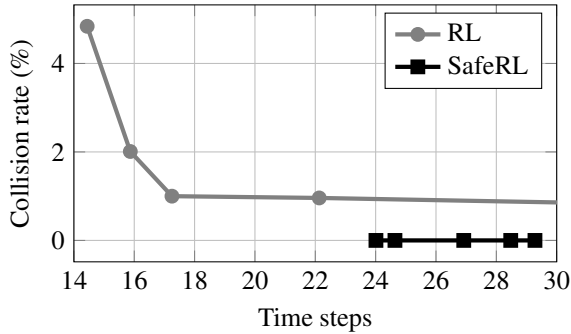

The problem of interest is a multi-objective optimization problem where the agent must balance two conflicting objective: safety and efficiency. We show the Pareto frontier for the safe RL and RL policies in \cref{fig:pareto}. As the reward functions parameters are varied we can generate different operating point in the safety versus efficiency diagram. In \cref{fig:pareto}, the region of optimality is in the bottom left corner. 

\subsection{Discussion}

In \cref{tab:results} we can see that all the policies relying on the model checking technique are safe. When taking random actions inside the set $Act(s)$ the agent can still navigate safely in the environment. This result provides an empirical proof that the safe RL approach will satisfy the specification even at early stage of training. Although no collisions occurred in ten thousand simulations, the probabilistic guarantee only ensures \num{0.9999} certainty on safety. With more simulations, collisions may occur. We believe that in our evaluation environment, this requirement is strong enough to prevent collisions.

The rule based method is dominated both in safety and efficiency by the safe RL and RL policy. 
Although the RL policy provides the fastest time to cross, it was not able to achieve the safety objective. 
Across the different collision penalty that we tried, \num{-1.5} gave the safest policy that did not result in an average time greater than the time achieved by safe random. Setting the collision cost to \num{-5} resulted in the ego vehicle staying at its initial position. 

As shown in \cref{fig:pareto}, the safe RL approach allows us to reach different operating points that are not accessible by tuning the reward function. The difficulty of trading off the two objectives by varying the weights in the reward function highlights the benefit of our approach. Since the safety objective is handled by the model checking step, the reward design is less challenging. In this problem, only efficiency needs to be encoded in the reward function for the safe RL policy.  

\Cref{fig:training} shows the performance of the policy throughout the training process for the scenario involving a car and a pedestrian. The safe RL policy did not collide during training, as expected from the performance of the safe random policy. Although the action space was constrained, the safe policy is able to accumulate as much reward as the non-safe RL agent. 

\section{Conclusion}

Reinforcement learning methods can derive policies in uncertain environments. Although the resulting behavior can be very efficient it is often challenging to draw guarantees on its performance. In this work we presented a two-step framework to compute policies satisfying probabilistic specification expressed as LTL formulas. The first step consists of identifying acceptable actions at each state via a model checking approach. The second step applies any RL algorithm to maximize an accumulated reward with a  modified exploration strategy. 
By constraining the agent to only choose acceptable actions, the probabilistic guarantee is transfered to the trained policy. 
Finally, we focused on applying the proposed approach to an autonomous driving scenario.

We demonstrated that our approach helps training significantly and simplifies the multi-objective optimization problem. 
Further work will extend the framework to partially observable domains. 
Such problems are especially relevant to driving scenarios where information about the intentions of other drivers is not directly accessible to the autonomous vehicle \citep{sunberg2017}. 
In addition, we will scale the model checking part of the framework to larger environments through local approximation strategies.  

\subsubsection*{Acknowledgement}
This work has been supported by the Wallenberg AI, Autonomous Systems and Software Program (WASP) and the Swedish Research Council (VR).

\subsubsection*{References}

\printbibliography[heading=none]

\end{document}

%% file: pareto.tex
\begin{tikzpicture}[]
\begin{axis}[legend pos = {north east}, ylabel = {Collision rate (\si{\percent})}, xlabel = {Time steps}, xmin = {14}, xmax = {30}, width=8cm, height=5cm, grid=both]\addplot+ [ultra thick, gray, mark=*, mark options={gray}]coordinates {
(14.44, 4.84)
(15.87, 2.01)
(17.25, 1.0)
(22.13, 0.96)
(96.0, 0.01)
};
\addlegendentry{RL}
\addplot+ [ultra thick, black, mark=square*, mark options={black}]coordinates {
(24.01, 0.0)
(24.64, 0.0)
(26.92, 0.0)
(28.477, 0.0)
(29.28, 0.0)
};
\addlegendentry{SafeRL}
\end{axis}

\end{tikzpicture}